# Open Knowledge Base Canonicalization with Multi-task Unlearning

Bingchen Liu, Shihao Hou, Weixin Zeng, Xiang Zhao, Shijun Liu, Li Pan

*Abstract*—The construction of large open knowledge bases (OKBs) is integral to many applications in the field of mobile computing. Noun phrases and relational phrases in OKBs often suffer from redundancy and ambiguity, which calls for the investigation on OKB canonicalization. However, in order to meet the requirements of some privacy protection regulations and to ensure the timeliness of the data, the canonicalized OKB often needs to remove some sensitive information or outdated data. The machine unlearning in OKB canonicalization is an excellent solution to the above problem. Current solutions address OKB canonicalization by devising advanced clustering algorithms and using knowledge graph embedding (KGE) to further facilitate the canonicalization process. Effective schemes are urgently needed to fully synergise machine unlearning with clustering and KGE learning. To this end, we put forward a multi-task unlearning framework, namely MulCanon, to tackle machine unlearning problem in OKB canonicalization. Specifically, the noise characteristics in the diffusion model are utilized to achieve the effect of machine unlearning for data in OKB. MulCanon unifies the learning objectives of diffusion model, KGE and clustering algorithms, and adopts a two-step multi-task learning paradigm for training. A thorough experimental study on popular OKB canonicalization datasets validates that MulCanon achieves advanced machine unlearning effects.

## I. Introduction

With the progress in closed information extraction (CIE) techniques [1], curated knowledge bases (CKBs) like YAGO [2] and Freebase [3] have developed rapidly. CKBs make a profound impact on various knowledge-driven applications in the field of mobile computing such as knowledge graph comparative reasoning systems [4] and semantic ranking for academic search [5]. However, new things and facts continue to emerge in the real world, which may not be covered by existing ontology for CIE, rendering CKBs inadequate to handle changes. As an alternative, open information extraction (OIE) can be leveraged [6], which extracts triples in the form of ⟨*head noun phrase*, *relational phrase*, *tail noun phrase*⟩ from the unstructured text; in this scenario, there is no need to designate an ontology in advance. The extracted triples together make a large open knowledge base (OKB), such as ReVerb [7]. While OKB comes with advantages over CKB in terms of coverage and diversity, there is a prominent issue associated with OKB that noun phrases (NPs) and relational phrases (RPs) in the triplets are not canonicalized, which leads to redundancy and ambiguity of knowledge facts.

In response, the task of OKB canonicalization is intensively studied and plays a significant role [8]. In the process of canonicalization, synonymous NPs (RPs) are aggregated into the same cluster, and thus the OIE triples can be converted into canonicalized forms. This greatly eliminates the redundant information and resolves the ambiguity of knowledge facts, providing a refined knowledge base for the downstream tasks. While some use the knowledge embeddings of OIE triples to conduct the clustering [9], some use the information in the source context to facilitate the clustering process [10]. Some fuse the two types of information sources in pursuit of a more accurate clustering result [11]. Unlike most of existing solutions that use the hard clustering strategy to assign each NP/RP to a single cluster, a recent work adopts variational autoencoder (VAE), a generative method, to tackle OKB canonicalization [12], which learns the embeddings and cluster assignments in an end-to-end fashion, resulting in better vector representations for NPs/RPs, and also state-of-the-art canonicalization results.

Nevertheless, some of the noun phrases in the formed OKBs extracted from the real world based on the OIE technique may become obsolete or lead to errors, resulting in the need for the involved OKB-related training models to be adapted over time and thus accordingly. For example, Lionel Messi joined Paris Saint-Germain Football Club in 2021 and Miami International Football Club in 2023, and his club information should be replaced over time. In addition, in OKBs involving areas such as healthcare, finance and insurance, there are privacy regulations involved, such as the EU General Data Protection Regulation (GDPR). At this point, there is a need to guarantee the right to be forgotten about the data. It is easy to remove this part of the data involved in privacy in OKB, but it is difficult to eliminate their impact on the trained embedding model. In addition, retraining the model again on the remaining data would take a lot of time and effort. At this point, machine unlearning as an excellent solution should be reasonably well embedded in the task of OKB canonicalization.

To address the above issues, we propose MulCanon, a novel framework for open knowledge base canonicalization with *multi-task unlearning*. In the learning process, the main

(*Corresponding author: Weixin Zeng*)

Bingchen Liu, Shijun Liu and Li Pan are with the School of Software, Shandong University, Jinan, 250098, China.(e-mail: lbcraf2018@126.com; lsj@sdu.edu.cn; panli@sdu.edu.cn )

Shihao Hou is with the School of Computer Science and Technology, Shandong University of Finance and Economics, Jinan, 250000, China.(e-mail: shsdufe@126.com )

Weixin Zeng and Xiang Zhao is with the College of Systems Engineering, National University of Defense Technology, Changsha, 410073, China.(e-mail: dexterdervish@foxmail.com; xiangzhao@nudt.edu.cn)

purpose is to assign NPs/RPs with the same meaning into the same cluster. During the unlearning process, remove a small portion of information related to privacy protection and outdated information, while ensuring that it does not affect the overall performance of the model. Firstly, we propose to use a new generative model, i.e., ***diffusion model***, in the soft clustering process [13]. Such a design allows for the clever use of noise in diffusion generation models to achieve oblivion removal for specific knowledge in the unlearning process. Secondly, we also consider KGE learning as a subtask, where the canonized NPs/RPs are required to meet the constraints induced by the KGE model. Thirdly, we integrate the learning objectives and use a ***two-step multi-task*** paradigm to train the model. Extensive experimental results validate that our proposal can consistently achieve superior machine unlearning effects on popular datasets.

The major contributions of this work are as follows:

- We use the diffusion model in the soft clustering process, which can effectively assist in forgetting special information in the unlearning process.
- We put forward MulCanon, a novel framework for machine unlearning in open knowledge base canonization. We integrate separated learning objectives and use a two-step multi-task paradigm for model training.
- Extensive experimental results on popular OKB canonicalization benchmarks demonstrate that MulCanon can achieve superior machine unlearning effects on popular datasets.

In Section II, we present the work related to OKB canonicalization and machine unlearning. In Section III, we give a formal definition of the problem. In section IV, we specify our proposed model and approach. In section V, we conduct experiments to verify the validity and soundness of the proposed approach. Finally, we conclude the text and look forward to the work ahead in section VI.

## II. RELATED WORK

In this section, we present the relevant work related to OKB canonicalization and machine unlearning. We divide the presentation into four aspects, OKB canonicalization, machine unlearning, diffusion model and entity alignment.

### A. OKB canonicalization

Research on the task side of OKB canonicalization can be divided into two types: semi-supervised methods [14] [15] [16] [17] and unsupervised methods [10] [11]. Regarding the unsupervised methods, Lin et al. [10] use relevant knowledge from the context of the extracted source text to aggregate NPs (RPs) into groups with similar respective interpretations. CMVC [11] in order to compensate for the shortcomings of using knowledge from only a single perspective, knowledge from both views is used in combination to obtain more accurate results. Unlike this article, the methods mentioned above do not require manual labelling for each group.

As to the semi-supervised methods, [14] applies hierarchical agglomerative clustering (HAC) algorithms to manually defined features such as overlapping IDF [14] tokens to group synonymous NPs into the same group, and then cluster RPs into the same group by using an association rule mining algorithm based on an incomplete knowledge base [18]. Wu et al. [17] use pruning and boundary techniques to reduce the similarity computation and build a graph-based clustering approach based on the previous canonicalization model [14], so as to improve the efficiency and applicability of the model. CESI [16] first learns the embedding of NPs (RPs) using a KGE model based on the factual view and and additionally using various edge information (e.g. like PPDB information [19], etc.), and performs the embedding based on the embedding to bring the interpretation of the same NPs (RPs) into a group. JOCL [15] explores a clustering scheme combining a canonicalization task and a linking task using a factor graph model. Like the MulCanon proposed in this paper, the model proposed by this type of model needs to label each generated group.

Unlike state-of-the-art approaches that focus on equating the OKB canonicalization problem to the clustering problem of NPs (RPs), ignoring the generation of more accurate representations of NPs (RPs), CUVA [12] applies variational autoencoders (VAE) to the process of learning NPs (RPs) representations. Nevertheless, the VAE generation process is not equidimensional, leading to distortion and warping of features.

### B. Machine unlearning

Machine unlearning is a different approach to traditional machine learning, where the main objective is to allow the model to actively forget or delete specific samples, thus enabling flexible management of learned information. Unlike retraining the entire model, the core idea of machine unlearning is to achieve forgetting by updating the sum of queries without losing overall performance, thus saving time and computational resources. A related work [20] has designed effective machine unlearning by using k-means clustering technique, which improves the deletion efficiency and enables the model to forget unwanted information more accurately and precisely. A study [21] introduces the concept of random forests and proposes an inert forgetting strategy that performs the operations involved in different suspension requests in a single batch to avoid redundant computations and achieve computation sharing. In the field of knowledge graph research, a completely new paradigm [22] has recently been proposed that employs a unlearning approach from cognitive neuroscience. This framework combines retrospective inhibition and automatic attenuation, aiming at removing specific knowledge from local customers and propagating this eliminated information into the global model by means of knowledge distillation. Despite this, few studies have explored unlearning work related to OKB, suggesting that the field of machine forgetting is still open to further research and innovation to meet evolving application needs.

### C. Diffusion model

The diffusion model defines a markov chain of diffusion steps to slowly add random noise to the data, and then learns the inverse diffusion process to construct the required data samples from the noise [23]. Over recent years diffusion models have achieved significant advantages in many areas such as anomaly detection [23], image generation [24] and image noise removal [25]. The diffusion model has significant operational advantages, which can efficiently capture the complex structure and underlying laws in the data, and performs better than the traditional discriminant model in the data generation task. The advantages of diffusion model such as efficient capture of data distribution, training using diffusion process, sample-level parallel computing, and wide application to different data types make it the focus of research and application in the field of generative models, and show great potential and application prospects in various fields.

To the best of our knowledge, this paper is the first work to use diffusion models to perform OKB canonicalization tasks. In this work, we also use the diffusion model in the OKB canonicalization process, which can avoid the distortion of data representations and lead to better clustering results.

### D. Entity alignment

The purpose of entity alignment is to determine whether two or more entities from different sources of information are pointing to the same object in the real world, and to bring together named entities that have the same referent object in different knowledge graphs. For instance, to alleviate the limitation of relying only on structural information, Zeng et al. [26] introduced entity name information and designed a common attention feature fusion network to adjust the weights of different features to improve the performance of the knowledge graph entity alignment. In addition, we also pay attention to the fact that many researchers have focused on entity alignment for temporal knowledge graphs in recent times, Liu et al. [27] proposed a box embedding for entity alignment of temporal knowledge graphs in the news domain. There are similarities between entity alignment for knowledge graphs and the OKB canonicalization task undertaken in this paper, and they are closely related. Nevertheless, OKB canonicalization involves the canonicalization of elements in a single knowledge graph, while entity alignment tends to take place among multiple KGs.

## III. Problem definition

In this section, we first give three definitions of relevant concepts. Then, we provide a formal definition of the problem in this paper.

**Definition 1** (Open Information Extraction (OIE)). *OIE targets at open texts and aims to extract a large number of relational tuples without any relation-specific training data. We give concrete examples of OIE in Figure 1.*

**Definition 2** (Open Knowledge Base (OKB)). *The OpenIE method performs extraction for a given unstructured text document under unsupervised conditions, and stores the extracted triadic forms (noun phrases, relational phrases, noun phrases) in a unified and centralized manner, thus forming a large OKB. Millions or even more of triples extracted by OIE are stored in OKB.*

Tons of OIE triples $< h, r, t >$ exist in OKB. The goal of the OKB canonicalization task is to use the triples and the information extracted from the source text $s$ to cluster synonymous NPs pointing to the same entity and synonymous RPs with the same semantics into a group, each of which converts these OIE triples into a canonical form.

**Definition 3** (Machine Unlearning). *Machine Unlearning refers to a research task in the field of machine learning and artificial intelligence whose main objective is to enable machine learning models to actively forget knowledge or information that has already been learnt. This is in contrast to traditional machine learning tasks, which typically involve improving model performance by continuously accumulating data and knowledge. The key idea of machine unlearning is to enable the model to selectively discard some or all of the learned knowledge in order to adapt to changing environments or new tasks while maintaining the overall performance of the model. This task is especially important for removing information that is outdated due to the passage of time as well as private information. For the example in Figure 1, Lionel Messi joined Paris Saint-Germain Football Club in 2021 and Miami International Football Club in 2023. Therefore, the "Paris Saint-Germain Football Club" involved in 2021 should be removed and forgotten because it's out of date.*

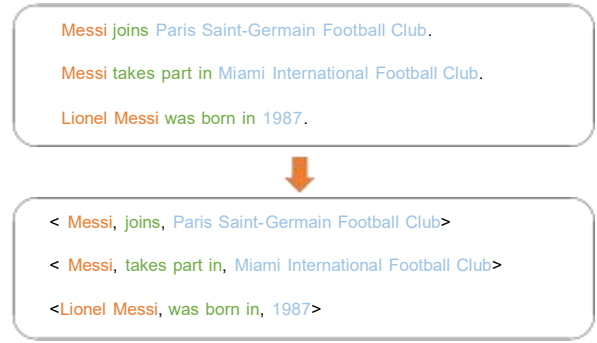

Fig. 1: An example of OIE.

**Example 1.** *As shown in Figure 1, there are three sentences concerning Messi, where an OIE system extracts three OIE triples. Unfortunately, it is not clear whether Messi and Lionel Messi refer to the same entity. This means that when the term Lionel Messi is queried in the above OIE triplet, the machine cannot fully query all available facts related to the entity. In addition, it can be shown that the first two OIE triples refer to the same meaning, and only one of them needs to be stored.*

## IV. The proposed method

In this section, we first describe the framework outline of our proposed model. Then, we present each part of the model

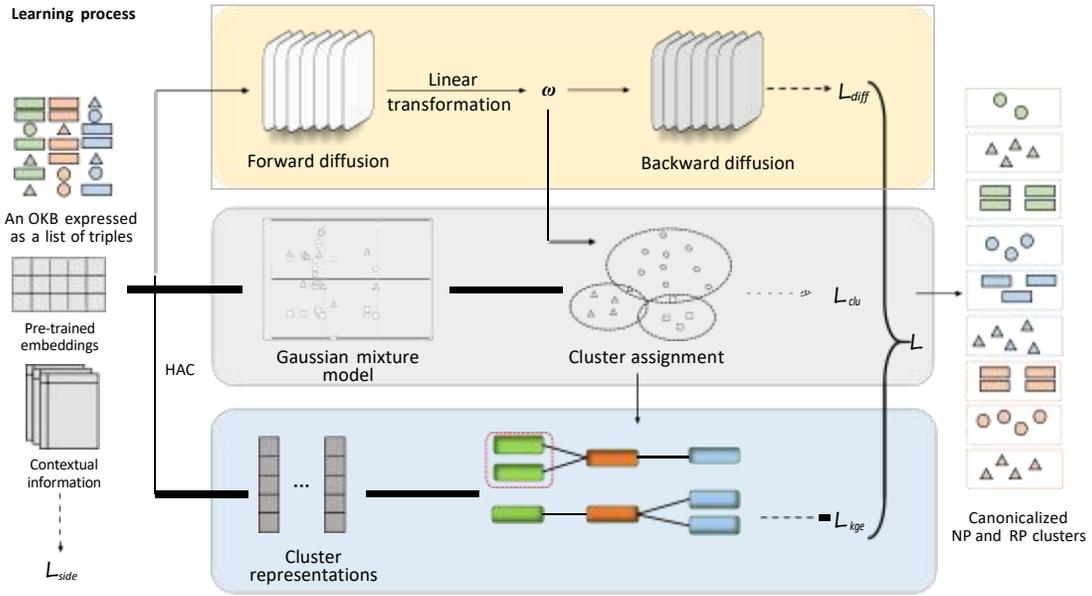

Fig. 2: The learning process of MulCanon.

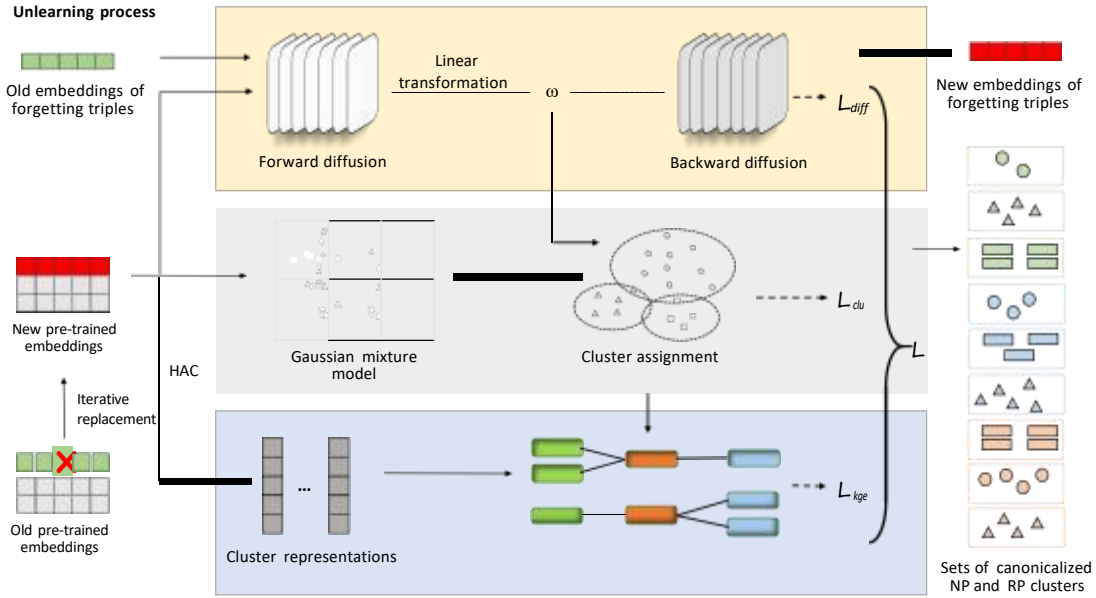

Fig. 3: The unlearning process of MulCanon.

separately. Finally, we present the training method and side information.

### A. Framework outline

The learning process of MulCanon is presented in Figure 2. The inputs include the OKB triples, consisting of head NPs (in green), RPs (in orange), and tail NPs (in blue), the contextual information of these triples, and pre-trained word embeddings such as GloVe [9]. We use Glove word embeddings here, in order to ensure consistency with the use of word embeddings across baseline models and thus ensure fairness in the comparison of experimental results.The goal is to generate the canonicalized forms of these phrases, as shown in the rightmost of Figure 2.

To this end, we devise a multi-task learning framework, MulCanon, to tackle OKB canonicalization. The main task is to divide the phrases into their corresponding cluster, and we use a generative framework, variational deep embedding (VaDE) [13], to conduct the soft clustering. Instead of adopting the original VAE model, we propose to utilize the diffusion model to avoid the potential information loss during the transformation, resulting in the diffusion loss $L_{diff}$. In the meantime, we use hierarchical agglomerative clustering (HAC) [28] upon the phrases to obtain an initial cluster assignment, which is also used to initialize the Gaussian mixture model. Then, based on the Gaussian mixture model and the middle variable **w** of the diffusion process, we predict the cluster assignment, which is contrasted with the clustering result produced by HAC, leading to the clustering loss $L_{clu}$. Note that the results produced by HAC are merely considered as *weak supervision*, and there is no actual supervision signal.

TABLE I: Important notations

| Notation | Description |
|---|---|
| $L_{diff}$ | diffusion loss |
| $L_{clu}$ | clustering loss |
| $L_{kge}$ | knowledge graph embedding loss |
| $L_{cl}$ | contrastive learning objective |
| $p(c)$ | prior probability of cluster assignment |
| $\mu_c$ | the mean of the mixed Gaussian distribution of cluster c |
| $\sigma_c^2$ | the variance of the mixed Gaussian distribution of cluster c |
| $\beta_t$ | variance schedule across diffusion steps |
| $t$ | number of steps |
| $z_1$ | noise sampled from the standard normal distribution |
| $x_{t-1}$ | the representation in the previous step of diffusion process |
| $T$ | step of diffusion process |
| $W$ | learnable matrice |
| $z$ | a noise that fits the standard normal distribution |
| $f_T(x_t, t)$ | a neural network to approximate the conditional probabilities |
| $L_1$ | the multi-task learning objective of step one |
| $L_2$ | the multi-task learning objective of step two |

As an additional task, KGE is conducted to ensure that the canonicalized NPs/RPs satisfy the inherent KG structure. Specifically, we use the predicted cluster assignment to obtain the representations for elements in OIE triples, and produce the KGE loss $L_{kge}$. Finally, we formulate the multi-task learning objective L and train the model, which is then used to infer the final canonicalization results.

The unlearning process of MulCanon is presented in Figure 3. After MulCanon has been trained and stabilised, the embedding of the data of the forgotten triples are made to pass through the diffusion model that has been trained in the sub-task, and through the diffusion model, new embeddings of the forgotten triples are re-generated and the newly generated embeddings and the retained set embeddings are re-combined to pass through the model. This process can make full use of the noise characteristics of the diffusion generation model, the feature information of the forgotten set is diluted and faded by noise in the diffusion generation model, so as to achieve the effect of targeted forgetting.

### B. Cluster assignment

Following existing literature [11], we consider OKB canonicalization as a clustering problem, and thus the main goal is to correctly predict the cluster assignment. The final sets of clusters $Set_1, ..., Set_n$, canonicalized after cluster assignment here, are used as the output of the model.

**HAC clustering.** We utilize the HAC clustering algorithm [29] to obtain a preliminary clustering result of the NPs and RPs, which takes the phrases and the pretrained embeddings as input. The embedding of the central node of each cluster is selected for initialising the Gaussian mixture model. During this process, we also generate the preliminary labels for NPs and RPs.

**Clustering prediction.** As shown in the center of Figure 2, in MulCanon, the probability of the actual cluster assignment is calculated via a softmax function, where the probability of assigning to cluster c is denoted as:

$$v(c) = \frac{p(c)p(w|c)}{\sum p(c)p(w|c)} \quad (1)$$

where $p(c)$ is the prior probability, calculated as the percentage of phrases belonging to cluster c in the Gaussian mixture model. $p(w|c)$ denotes the probability of sampling **w** from $VI(\mu_c, \sigma_c^2)$, where $\mu_c$ and $\sigma_c^2$ are the mean and variance of the mixed Gaussian distribution corresponding to the cluster c. **w** is obtained by the diffusion model process, which will be detailed in the next subsection. Then, we consider the label generated by the HAC algorithm as weak supervision, and contrast it with the predicted result **v** using the cross-entropy cost function, leading to the clustering loss $L_{clu}$. Notably, the labels produced by the HAC algorithm are only weak supervision, and the potential errors will be corrected by the multi-task learning process. MulCanon is still an unsupervised framework that does not require labeled data.

### C. Diffusion model

We adopt a generative framework, VaDE [13], to conduct the soft clustering, which generates **w** and forwards to the clustering assignment process. Specifically, we propose to utilize the diffusion model that performs equidimensional transformation, instead of the original VAE model used in [13].

**Forward diffusion process.** The forward diffusion process is a continuous noise addition process, calculated as follows:

$$x_t = \sqrt{\alpha_t} x_{t-1} + \sqrt{1 - \alpha_t} z_1 \quad (2)$$
$$\alpha_t = 1 - \beta_t \quad (3)$$

where $\beta_t$ represents the variance schedule across the diffusion steps [23]. t is the number of steps, $Z_1$ is the noise sampled from the standard normal distribution, $Z_{t-1}$ is the representation in the previous step, and $Z_0$ in the first step is initialized with the GloVe embedding of the corresponding NP or RP.

After obtaining the representations in the last step T, $Z_T$, we use the reparametrization trick to sample **w**:

$$w = \mu + \sigma * Z_0 \quad (4)$$

where $\mu = x_T W_\mu$ and $\sigma = x_T W_\sigma$. $W_\mu$ and $W_\sigma$ are two learnable matrices, and $Z_0$ is a noise that fits the standard normal distribution.

**Reverse diffusion process.** Then we conduct the reverse diffusion, which progressively removes noise, starting with a random noise and gradually reducing to a representation without noise:

$$\mathbf{x}_{t-1} = \frac{1}{\sqrt{\alpha_t}} \mathbf{x}_t - \frac{\sqrt{1-\alpha_t}}{\sqrt{\alpha_t}} f_\tau(\mathbf{x}_t, t) + \mathbf{z}_2 \quad (5)$$

where $\alpha_t$ is calculated using Formula 3, $\mathbf{z}_2$ is also a noise that fits the standard normal distribution, $f_\tau(\mathbf{x}_t, t)$ is a neural network to approximate the conditional probabilities, which is implemented by multilayer perceptron, and $\tau$ refers to the trainable parameter in the network.

**Diffusion loss.** The diffusion process aims to train $f_\tau$ so that its predicted noise is similar to the real noise used for destruction. Following existing literature [23], the diffusion loss $L_{diff}$ is:

$$L_{diff} = \mathbb{E}_{t\sim[0-T], f\sim\mathcal{N}(0,I)} [||f - f_\tau(\mathbf{x}_t, t)||^2] \quad (6)$$

where $f$ is sampled from a standard Gaussian distribution, $t$ denotes the time step, and $||\cdot||$ denotes the L2 distance.

Compared with the hard clustering strategy used in the majority of existing works, the soft clustering of the generative model can help explain different meanings of a given entity mention. This advantage reasonably compensates for the shortcomings of hard clustering methods that only assign each entity to a cluster. That being said, traditional generative models such as generative adversarial network (GAN) and VAE involve dimensional compression and expansion, which might lead to information distortion and loss during data transformation. Hence, in this work, we fill in this gap by using the diffusion model.

### D. Side information

With reference to [16], we describe in this section several types of side information used in this work. The following four sources of side information are used, as described below:

- PPDB information: PPDB 2.0 [30] is used here to determine whether two NPs are equivalent to each other. This is done by extracting high-confidence paraphrases from the dataset and removing overlapping redundancies in the first step. In the second step, the information extracted in the previous step is found in a concatenated set, all equivalent NPs are clustered, and a representation is defined for each cluster. In the third step, a search is performed to find out whether two NPs have the same representation and thus determine the equivalence between them.

- Entity linking: Compared to the process of PPDB information, this information is relatively simple to obtain. To obtain entity linking information, it is only necessary to use the entity linker [31] for a given text to map the noun phrases (NPs) to the entities present in Wikipedia. The equivalence between them is determined based on whether they are linked to the same entity or not.

- Morphological canonicalization: we find equivalent NPs using the morphological canonicalization method used in open information extraction [32].

- IDF token overlap: the rationale for the use of this information is that if NPs or RPs share infrequently used terms, they are considered to have a higher probability of referring to the same entity (or relationship), which is considered a valid feature for use. For the specific formula for calculating overlap scores we refer to the formula provided in [16], and we remove pairs with scores less than a specific threshold.

### E. Multi-task learning objective

In addition to the clustering assignment objective with diffusion model and the KGE learning objective, we also adopt the side information objective $L_{side}$ by following [12] and [16]. The main motivation is to harness the contextual information to further facilitate the overall learning process.

As to the overall training procedure, we adopt a two-step strategy. In step one, the multi-task learning objective is:

$$L_1 = L_{clu} + L_{diff} + L_{side} \quad (7)$$

In step two, the multi-task learning objective is:

$$L_2 = L_{clu} + L_{diff} + L_{kge} + L_{side} \quad (8)$$

Noteworthily, by conducting the multi-task learning, Mul-Canon can better capture the interactions among sub-tasks, which can further regulate the canonicalization process and leading to better results. Besides, the motivation for adopting the two-step training strategy is that training KGE requires the outputs of clustering. Thus, by training the generative clustering process first, we can obtain relatively stable parameters of the clustering module, based on which the contrastive KGE can be better trained.

## V. EXPERIMENTS

In this section, we conduct experiments, thus verifying the validity and reasonableness of proposed model.

### A. Experimental setting

**Datasets.** Three datasets are used:
- ReVerb45K [16] is the benchmark dataset commonly used for evaluating OKB canonicalization tasks. The OIE triples of ReVerb45K are extracted by ReVerb [7] from the source text of Clueweb09 [33], and the NPs are annotated with the corresponding entities in Freebase. The number of triples in ReVerb45K is 45K, each triple of which is associated with a Freebase entity. For each entity, there are two or more aliases in the NPs.

**Algorithm 1** The main process of Mulcanon

**Inputs:**
$OKB_1 = (h_{n_1}, r_{n_1}, t_{n_1})$: an OKB expressed as a list of triples
$S_n$: the contextual information of the above triples
**Output:**
$Set_1, ..., Set_n$: Sets of canonicalized clusters

1: **for** i = 1, ..., n **do**
2:    Generate a preliminary clustering results of $OKB_1$ and preliminary labels by HAC clustering algorithm
3: **end for**
4: **for** i = 1, ..., n **do**
5:    Generate the representation $x_0$ with the Glove embedding
6: **end for**
7: **for** i = 1, ..., n **do**
8:    Calculate $x_t$ using Eq.(2)-Eq.(3)
9:    Calculate $\omega$ using Eq.(4)
10:    **for** i = 1, ..., n **do**
11:       Calculate the probability of assigning to cluster c using Eq.(1)
12:       Calculate $x_{t-1}$ using Eq.(5) and get $x_0$
13:       Optimization using Eq.(6) and Eq.(7)
14:    **end for**
15: **end for**
16: **for** i = 1, ..., n **do**
17:    Optimization using Eq.(6) and Eq.(8)
18: **end for**
19: **for** i = 1, ..., k **do**
20:    Calculate $\check{x}_t$ using Eq.(2)-Eq.(3)
21:    Calculate $\omega$ using Eq.(4)
22:    Calculate $\check{x}_{t-1}$ using Eq.(5) and get $\check{x}_0$
23: **end for**
24: Update $x_0$ based on $\check{x}_0$
25: repeat lines 1 to 18
26: Calculate evaluation index
27: **return** $Set_1, ..., Set_n$

- Canonicnell [12] The data in this dataset was collected from [34] generated data artefacts using an automated construction strategy, from which the cumulative knowledge was extracted as a list of (topic, relation, object) triples to form this dataset. This is a newly proposed dataset in which the co-referenced entities of the triples are labelled, pairs with scores not greater than 0.25 are filtered, and the filtered remaining pairs are considered as undirected edges in the graph. A depth-first search strategy is used to obtain a set of connected components, thus constituting the set of gold clusters, for which entities appearing as heads and tails are retained, and in this way the remaining list in Canonicnell is filtered out. The triples obtained from the above process are merged with the previously obtained gold clusters to form a new classical dataset.

Details of the datasets are summarized in Table II. 20 percent of the triples are used for validation, and the rest of triples are regarded as the test set. The validation set is used to tune the hyper-parameters and there are no labeled data.

TABLE II: Details of datasets used.

| Datasets | Gold entities | NPs | RPs | Triples |
|---|---|---|---|---|
| Reverb45k | 7.5K | 15.5K | 22K | 45K |
| Canonicnell | 1.4K | 8.7K | 139 | 20K |

**Implementation details.** As introduced in Section IV-E, we adopt a two-step training paradigm. In the first stage, we mainly train the clustering and diffusion models, and do not include KGE training. The learning rate is set to $10^{-3}$. In the second stage, we train KGE. The learning rate is set to $10^{-5}$. The number of training rounds for each batch is set to 50. In the input side, if an NP contains two or more words, its embedding is defined as the average of the embeddings of the individual words contained in it. As to the diffusion model, we set the time step T to 2, which will be further analyzed in the hyper-parameter analysis. For the loss function associated with the HolE algorithm, we randomly select 20 negative samples for each positive sample. In addition, experiments are conducted using one Intel x86 CPU and one NVIDIA GeForce RTX 3090 GPU with a maximum of 24GB RAM, implemented in Python 3.9.0.

**Evaluation metrics.** We adopt the F1 metric for evaluating entity canonicalization, which is the harmonic mean of precision (P) and recall (R). Following previous works [16] [14], we report results of three variants, i.e., macro F1, micro F1 and pair F1. Specifically,

- Macro. P is defined as the fraction of pure clusters in C, that is, all NPs (RPs) are linked to clusters of the same gold entity (or relationship). The calculation of R is similar to that of P, but the functions of E and C are interchangeable.
- Micro. P is defined as cluster C is based on the assumption that the most frequent gold entities (or relationships) in the cluster are correct. The definition of R is similar to the macro R.
- Pair. P is measured by the ratio of the number of hits in C to the total possible matches in C. R is the ratio of hits in C to all possible pairs in E. If a pair of elements in the cluster in C all refer to the same gold entity (or relationship), a hit will occur.

More details are omitted in the interest of space. [16] is recommended for interested authors. Notably, following [12], we do not report the results on relation canonicalization, as benchmarks before do not contain ground-truth annotations for relations.

### B. Main experiments

**Baselines.** Referring to the evaluation approach of recent research [35] on machine unlearning in knowledge graph related areas, we used several sets of controlled experiments to validate the effectiveness and feasibility of the proposed unlearning process. In the main experiment phase, three sets of raw, retrained and unlearned embeddings are used to be compared. The raw embeddings are derived from embeddings trained by the original full set of data. The re-trained embeddings are derived from embeddings trained using the remaining retained set of data. The unlearned embeddings are derived from embeddings trained on the unlearning process based on the raw embeddings. The main experimental phase of the main experiment is the comparison of the three sets of data.

TABLE III: Unlearning results on Reverb45k datasets

| Macro F1 | Raw | | Re-trained | | Unlearn | |
|---|---|---|---|---|---|---|
| | Forget↓ | Test↑ | Forget↓ | Test↑ | Forget↓ | Test↑ |
| TransE | 0.9379 | 0.7481 | 0.9288 | 0.7182 | 0.6291 | 0.7415 |
| Hole | 0.9379 | 0.7481 | 0.9288 | 0.7023 | 0.6711 | 0.7462 |
| Micro F1 | Raw | | Re-trained | | Unlearn | |
| | Forget↓ | Test↑ | Forget↓ | Test↑ | Forget↓ | Test↑ |
| TransE | 0.9493 | 0.8316 | 0.9452 | 0.8215 | 0.5282 | 0.8313 |
| Hole | 0.9493 | 0.8318 | 0.9452 | 0.8118 | 0.5358 | 0.8285 |
| Pair F1 | Raw | | Re-trained | | Unlearn | |
| | Forget↓ | Test↑ | Forget↓ | Test↑ | Forget↓ | Test↑ |
| TransE | 0.8184 | 0.7938 | 0.8167 | 0.7892 | 0.1578 | 0.7937 |
| Hole | 0.8184 | 0.7935 | 0.8167 | 0.7882 | 0.1516 | 0.7902 |
| Average F1 | Raw | | Re-trained | | Unlearn | |
| | Forget↓ | Test↑ | Forget↓ | Test↑ | Forget↓ | Test↑ |
| TransE | 0.9019 | 0.7917 | 0.8969 | 0.7784 | 0.4384 | 0.7888 |
| Hole | 0.9019 | 0.7911 | 0.8969 | 0.7668 | 0.4528 | 0.7883 |

TABLE IV: Unlearning results on Canonicnell datasets

| Macro F1 | Raw | | Re-trained | | Unlearn | |
|---|---|---|---|---|---|---|
| | Forget↓ | Test↑ | Forget↓ | Test↑ | Forget↓ | Test↑ |
| TransE | 0.9676 | 0.7547 | 0.9865 | 0.6542 | 0.6351 | 0.7570 |
| Hole | 0.9897 | 0.7542 | 0.9865 | 0.6542 | 0.9646 | 0.7568 |
| Micro F1 | Raw | | Re-trained | | Unlearn | |
| | Forget↓ | Test↑ | Forget↓ | Test↑ | Forget↓ | Test↑ |
| TransE | 0.9270 | 0.8147 | 0.9867 | 0.7411 | 0.5372 | 0.8173 |
| Hole | 0.9869 | 0.8144 | 0.9867 | 0.7411 | 0.9663 | 0.8173 |
| Pair F1 | Raw | | Re-trained | | Unlearn | |
| | Forget↓ | Test↑ | Forget↓ | Test↑ | Forget↓ | Test↑ |
| TransE | 0.9328 | 0.2799 | 0.9214 | 0.2604 | 0.1711 | 0.3204 |
| Hole | 0.9328 | 0.2818 | 0.9214 | 0.2604 | 0.1711 | 0.3209 |
| Average F1 | Raw | | Re-trained | | Unlearn | |
| | Forget↓ | Test↑ | Forget↓ | Test↑ | Forget↓ | Test↑ |
| TransE | 0.6742 | 0.6164 | 0.9549 | 0.5519 | 0.4478 | 0.6316 |
| Hole | 0.9843 | 0.2538 | 0.6168 | 0.5519 | 0.7007 | 0.6317 |

A sample of 3% of the original set is generated as the forget set, and the remaining set is retained. According to the baseline setting method, the experiment will verify the evaluation effect of the model on the forgotten set and the original full set, where "forget" is the forgotten set and "test" is the full set in Table III and Table IV. In the Table III and Table IV, "forget" is the forgotten set and "test" is the full set. The lower the performance of the three selected evaluation metrics on the forgotten set, the more complete the unlearning effect on the forgotten set. We used the "test" full set to verify the cost of forgetting that had to be sacrificed. The experimental metrics for Macro F1, Micro F1 and Pair F1 in all main experiments are shown in Table III.

**Results.** From the experimental results in Table III, the following conclusions can be drawn: (1) The Macro F1, Micro F1 and Pair F1 scores of the retrained model on the "forget" set are very close to the scores on the "test" set. This suggests that the retrained model is still able to predict the missing links, i.e., complete forgetting cannot be achieved by retraining based on direct deletion of the training data. (2) The scores of the unlearned embeddings and the retrained embeddings are lower than the level of the original embeddings, suggesting that MulCanon is able to completely remove knowledge that needs to be forgotten and achieve memory activation on the unlearned set. (3) The learned embeddings have comparable performance to the original model on the test set, which is higher than the retrained model. This suggests that forgetting of MulCanon can be maintained without performance degradation for the whole model, and is superior to the approach of directly removing forgotten data for retraining.

TABLE V: Performance of the NP canonicalization task in ReVerb45K.

|  | Macro F1 | Micro F1 | Pair F1 | Average F1 |
|---|---|---|---|---|
| Morph Norm | 0.627 | 0.558 | 0.334 | 0.506 |
| Text Similarity | 0.625 | 0.566 | 0.394 | 0.528 |
| IDF | 0.603 | 0.551 | 0.338 | 0.497 |
| Attribute Overlap | 0.621 | 0.558 | 0.342 | 0.507 |
| CESI | 0.640 | 0.855 | 0.842 | 0.779 |
| JOCL | 0.537 | 0.854 | 0.823 | 0.738 |
| MulCanon | 0.751 | 0.833 | 0.795 | **0.793** |

### C. Comparison of learning process of MulCanon

**Methods for comparison in the learning process.** In order to verify that the performance of MulCanon during the learning process is comparable to the performance of other advanced OKB canonicalization models, we compared the performance of some other methods. We use the following methods for comparison:

- IDF [14], which uses the inverse document frequency (IDF) of the tokens in two NPs to calculate similarity and cluster them using HAC.

- Morph Norm [9], which divides NPs into groups after unifying grammars, e.g., tense.

- Text Similarity [10], which uses the Jaro-Winkler similarity [36] as a criterion for discriminating the similarity of NPs and clusters them using HAC.

- Attribute Overlap [15], which uses Jaccard similarity as a criterion for determining the similarity of NPs and clusters them using HAC.

- CESI [16], which fuses the knowledge embeddings of OIE triples and side information in pursuit of a more accurate clustering result.

- JOCL [15], which uses a factor graph model to combine the OKB canonicalization and entity linking tasks, mutually reinforcing the efficiency of both.

Table V report the NP canonicalization results on ReVerb45K. Referring to the experimental evaluation approach of the previous work on OKB canonicalization, we use the average F1 as primary reference for comparing the performance of models. It can be seen that our proposal MulCanon model outperforms all existing models in terms of *average F1* on all datasets. Next, we analyze the results in detail.

**Traditional solutions.** Specifically, the results of the heuristics, i.e., IDF and Morph Norm, are not satisfying, attaining merely 0.497 and 0.506 of average F1 on ReVerb45K, respectively. Both Text Similarity and Attribute Overlap use the HAC method for clustering. The difference between the two is that Text Similarity performs clustering based on the similarity between the contextual texts of the two NPs, whereas Attribute Overlap considers the similarity between the attributes of the two NPs. Still, compared with IDF and Morph Norm, they only slightly improve the performance. The inferior results of these simple methods can be ascribed to the fact that they mainly focus on leveraging the surface form variations of NPs while largely neglecting the other information, e.g., structural information in KG.

**Recent advanced solutions.** The recent advanced models largely improve the results. Specifically, CESI fully exploits the side information to facilitate the canonicalization process, which greatly improves the accuracy of clustering. The employment of side information has also been exploited by most of the recent methods. While JOCL considers the task of linking entities for joint learning, its performance is limited by its lack of exploitation of factors other than string and word embedding similarity of NPs.

**Our proposal.** During the learning process, different from all baseline models, MulCanon adopts a multi-task learning framework, which can fully exploit the synergy among clustering, diffusion, KGE learning and side information modeling to tackle the canonicalization task. Particularly, the diffusion model can help ensure iso-dimensional vector representations during data transformation process and avoid potential information loss.

For these reasons, MulCanon is competitive in terms of the performance of the learning process compared to the current state-of-the-art of its model.

### D. Hyper-parameter Analysis

**Different proportions of forgetting sets.** To analyze the influence of the proportions of forgetting sets on the effectiveness of MulCanon, we conduct the parameter analysis. Specifically, we set the proportions of forgetting sets from 2% to 5% , respectively, and report the average F1 results in Table VI. It observes that the proportions of forgetting sets do affect the results, and the performance decreases given more proportions, as larger proportion of the forgetting set leads to a more pronounced manifestation of eventual forgetting. In the main experiment, we randomly set the proportions of forgetting sets to a relative middle value.

## VI. CONCLUSION AND FUTURE WORK

This paper proposes a multi-task unlearning framework (MulCanon) for open knowledge base canonicalization, and introduces the diffusion model to further facilitate this process. Extensive experimental results on popular datasets demonstrate that our proposed model achieves advanced machine unlearning performance on the OKB canonicalization task.

TABLE VI: Unlearning performance with different proportions of forgetting sets in ReVerb45K.

|  | Macro F1 | Micro F1 | Pair F1 | Average F1 |
|---|---|---|---|---|
| 2% | 0.6974 | 0.6877 | 0.4499 | 0.6117 |
| 3% | 0.6532 | 0.6604 | 0.3725 | 0.5620 |
| 4% | 0.6414 | 0.6361 | 0.3565 | 0.5447 |
| 5% | 0.5810 | 0.4969 | 0.1168 | 0.3982 |

In future work, we will explore combining more subtasks to better complete the machine unlearning tasks of OKB canonicalization. In addition, entity linking and other forms of side information will also be the direction of innovation that we will try to experiment within the future. In addition, we believe that the machine unlearning tasks of some open knowledge bases or open knowledge graphs with temporal information will be interesting directions to explore in the future.

ACKNOWLEDGMENTS

The authors would like to acknowledge the support provided by the Innovation Method Project of Ministry of Science and Technology, China under Grant 2020IM020100, the Key Research and Development Program of Shandong Province (2020CXGC010102), project ZR2020LZH011 supported by Shandong Provincial Natural Science Foundation.

REFERENCES

[1] C.-H. Chang, M. Kayed, M. Girgis, and K. Shaalan, "A survey of web information extraction systems," IEEE Transactions on Knowledge and Data Engineering, vol. 18, no. 10, pp. 1411–1428, 2006. [Online]. Available: http://ieeexplore.ieee.org/document/1683775/

[2] F. M. Suchanek, G. Kasneci, and G. Weikum, "YAGO: A core of semantic knowledge unifying WordNet and wikipedia," in Proceedings of the 2007 World Wide Web Conference on World Wide Web - WWW '07, 2007, p. 449–458. [Online]. Available: https://hal.archives-ouvertes.fr/hal-01472497

[3] K. Bollacker, C. Evans, P. Paritosh, T. Sturge, and J. Taylor, "Freebase: A shared database of structured general human knowledge," in Proceedings of the Special Interest Group on Management Of Data-SIGMOD'08, 2008, p. 1247–1250. [Online]. Available: https://dl.acm.org/doi/10.5555/1619797.1619981

[4] L. Liu, B. Du, H. Ji, and H. Tong, "KompaRe: A knowledge graph comparative reasoning system," in Proceedings of the 27th ACM SIGKDD Conference on Knowledge Discovery and Data Mining, 2020, pp. 3308–3318. [Online]. Available: http://arxiv.org/abs/2011.03189

[5] C. Xiong, R. Power, and J. Callan, "Explicit semantic ranking for academic search via knowledge graph embedding," in Proceedings of the 26th International Conference on World Wide Web, 2017, pp. 1271–1279. [Online]. Available: https://dl.acm.org/doi/10.1145/3038912.3052558

[6] G. Angeli, M. J. Johnson Premkumar, and C. D. Manning, "Leveraging linguistic structure for open domain information extraction," in Proceedings of the 53rd Annual Meeting of the Association for Computational Linguistics and the 7th International Joint Conference on Natural Language Processing (Volume 1: Long Papers), 2015, pp. 344–354. [Online]. Available: http://aclweb.org/anthology/P15-1034

[7] A. Fader, S. Soderland, and O. Etzioni, "Identifying relations for open information extraction," in In Proceedings of the 2011 Conference on Empirical Methods in Natural Language Processing, 2011, p. 1535–1545. [Online]. Available: https://aclanthology.org/D11-1142

[8] D. Sturgeon, "Constructing a crowdsourced linked open knowledge base of chinese history," in 2021 Pacific Neighborhood Consortium Annual Conference and Joint Meetings (PNC), 2021, pp. 1–6. [Online]. Available: https://ieeexplore.ieee.org/document/9672294/

[9] J. Pennington, R. Socher, and C. Manning, "Glove: Global vectors for word representation," in Proceedings of the 2014 Conference on Empirical Methods in Natural Language Processing (EMNLP), 2014, pp. 1532–1543. [Online]. Available: http://aclweb.org/anthology/D14-1162

[10] X. Lin and L. Chen, "Canonicalization of open knowledge bases with side information from the source text," in 2019 IEEE 35th International Conference on Data Engineering (ICDE), 2019, pp. 950–961. [Online]. Available: https://ieeexplore.ieee.org/document/8731346/

[11] W. Shen, Y. Yang, and Y. Liu, "Multi-view clustering for open knowledge base canonicalization," in Proceedings of the 28th ACM SIGKDD Conference on Knowledge Discovery and Data Mining, 2022, pp. 1578–1588. [Online]. Available: https://dl.acm.org/doi/10.1145/3534678.3539449

[12] S. Dash, G. Rossiello, N. Mihindukulasooriya, S. Bagchi, and A. Gliozzo, "Open knowledge graphs canonicalization using variational autoencoders," in Proceedings of the 2021 Conference on Empirical Methods in Natural Language Processing (EMNLP), 2021, p. 10379–10394. [Online]. Available: http://arxiv.org/abs/2012.04780

[13] Z. Jiang, Y. Zheng, H. Tan, B. Tang, and H. Zhou, "Variational deep embedding: An unsupervised and generative approach to clustering," 2017. [Online]. Available: http://arxiv.org/abs/1611.05148

[14] L. Galárraga, G. Heitz, K. Murphy, and F. M. Suchanek, "Canonicalizing open knowledge bases," in Proceedings of the 23rd ACM International Conference on Conference on Information and Knowledge Management, 2014, pp. 1679–1688. [Online]. Available: https://dl.acm.org/doi/10.1145/2661829.2662073

[15] Y. Liu, W. Shen, Y. Wang, J. Wang, Z. Yang, and X. Yuan, "Joint open knowledge base canonicalization and linking," in Proceedings of the 2021 International Conference on Management of Data, 2021, pp. 2253–2261.

[16] S. Vashishth, P. Jain, and P. Talukdar, "CESI: Canonicalizing open knowledge bases using embeddings and side information," in Proceedings of the 2018 World Wide Web Conference on World Wide Web - WWW '18, 2018, pp. 1317–1327. [Online]. Available: http://arxiv.org/abs/1902.00172

[17] T.-H. Wu, Z. Wu, B. Kao, and P. Yin, "Towards practical open knowledge base canonicalization," in Proceedings of the 27th ACM International Conference on Information and Knowledge Management, 2018, pp. 883–892. [Online]. Available: https://dl.acm.org/doi/10.1145/3269206.3271707

[18] L. A. Galárraga, C. Teflioudi, K. Hose, and F. Suchanek, "AMIE: association rule mining under incomplete evidence in ontological knowledge bases," in Proceedings of the 22nd international conference on World Wide Web - WWW '13, 2013, pp. 413–422. [Online]. Available: http://dl.acm.org/citation.cfm?doid=2488388.2488425

[19] E. Pavlick, P. Rastogi, J. Ganitkevitch, B. Van Durme, and C. Callison-Burch, "PPDB 2.0: Better paraphrase ranking, fine-grained entailment relations, word embeddings, and style classification," in Proceedings of the 53rd Annual Meeting of the Association for Computational Linguistics and the 7th International Joint Conference on Natural Language Processing (Volume 2: Short Papers), 2015, pp. 425–430. [Online]. Available: http://aclweb.org/anthology/P15-2070

[20] G. V. Antonio Ginart, Melody Y. Guan and J. Zou, "Making ai forget you: Data deletion in machine learning," in Proceedings of the 33rd International Conference on Neural Information Processing Systems, 2019, p. 3513–3526.

[21] N. Sun, N. Wang, Z. Wang, J. Nie, Z. Wei, P. Liu, X. Wang, and H. Qu, "Lazy machine unlearning strategy for random forests," in Web Information Systems and Applications: 20th International Conference. Berlin, Heidelberg: Springer-Verlag, 2023, p. 383–390. [Online]. Available: https://doi.org/10.1007/978-981-99-6222-8_32

[22] X. Zhu, G. pu Li, and W. Hu, "Heterogeneous federated knowledge graph embedding learning and unlearning," in Proceedings of the ACM Web Conference 2023, 2023. [Online]. Available: https://api.semanticscholar.org/CorpusID:256615275

[23] J. Wyatt, A. Leach, S. M. Schmon, and C. G. Willcocks, "AnoDDPM: Anomaly detection with denoising diffusion probabilistic models using simplex noise," in 2022 IEEE/CVF Conference on Computer Vision and Pattern Recognition Workshops (CVPRW), 2022, pp. 649–655. [Online]. Available: https://ieeexplore.ieee.org/document/9857019/

[24] S. Gu, D. Chen, J. Bao, F. Wen, B. Zhang, D. Chen, L. Yuan, and B. Guo, "Vector quantized diffusion model for text-to-image synthesis," in 2022 IEEE/CVF Conference on Computer Vision and Pattern Recognition Workshops (CVPR), 2022. [Online]. Available: https://arxiv.org/abs/2111.14822

[25] X. Shan, J. Sun, Z. Guo, W. Yao, and Z. Zhou, "Fractional-order diffusion model for multiplicative noise removal in texture-rich images and its fast explicit diffusion solving," BIT Numerical Mathematics, vol. 62, no. 4, pp. 1319–1354, 2022. [Online]. Available: https://link.springer.com/10.1007/s10543-022-00913-3

[26] W. Zeng, X. Zhao, W. Wang, J. Tang, and Z. Tan, "Degree-aware alignment for entities in tail," in Proceedings of the 43rd International

*ACM SIGIR Conference on Research and Development in Information Retrieval*, ser. SIGIR '20. New York, NY, USA: Association for Computing Machinery, 2020, p. 811–820. [Online]. Available: https://doi.org/10.1145/3397271.3401161

[27] "Enhancing temporal knowledge graph alignment in news domain with box embedding, author = B. Liu et al., langid = english, journal = "ieee transactions on computational social systems"."

[28] P.-N. Tan, M. Steinbach, and V. Kumar, "Introduction to data mining, (first edition)," *Addison-Wesley Longman Publishing Co., Inc., Boston, MA, USA.*, 2005.

[29] M. Surdeanu, J. Tibshirani, R. Nallapati, and C. D. Manning, "Multi-instance multi-label learning for relation extraction," 2012. [Online]. Available: https://dl.acm.org/doi/10.5555/2390948.2391003

[30] E. Pavlick, P. Rastogi, J. Ganitkevitch, B. Van Durme, and C. Callison-Burch, "PPDB 2.0: Better paraphrase ranking, fine-grained entailment relations, word embeddings, and style classification," in *Proceedings of the 53rd Annual Meeting of the Association for Computational Linguistics and the 7th International Joint Conference on Natural Language Processing (Volume 2: Short Papers)*, pp. 425–430. [Online]. Available: http://aclweb.org/anthology/P15-2070

[31] M. Schmitz, R. Bart, S. Soderland, and O. Etzioni, "Open language learning for information extraction."

[32] A. Fader, S. Soderland, and O. Etzioni, "Identifying relations for open information extraction."

[33] M. Smucker, C. Clarke, and G. Cormack, "Experiments with clueweb09: Relevance feedback and web tracks." 01 2009. [Online]. Available: https://www.researchgate.net/publication/221038320_Experiments_with_ClueWeb09_Relevance_Feedback_and_Web_Tracks

[34] M. H. G. L. Pujara, Jay and W. Cohen, "Knowledge graph identification," in *The Semantic Web – ISWC 2013*. Berlin, Heidelberg: Springer Berlin Heidelberg, 2013, pp. 542–557.

[35] X. Zhu, G. Li, and W. Hu, "Heterogeneous federated knowledge graph embedding learning and unlearning," ser. WWW '23. New York, NY, USA: Association for Computing Machinery, 2023. [Online]. Available: https://doi.org/10.1145/3543507.3583305

[36] W. Winkler, "The state of record linkage and current research problems," *Statist. Med.*, vol. 14, 10 1999. [Online]. Available: https://www.researchgate.net/publication/2509449_The_State_of_Record_Linkage_and_Current_Research_Problems